\documentclass[pdflatex,sn-mathphys-num]{sn-jnl}% Math and Physical Sciences Numbered Reference Style 
\usepackage{natbib}
\usepackage{graphicx}%
\usepackage{multirow}%
\usepackage{amsmath,amssymb,amsfonts}%
\usepackage{amsthm}%
\usepackage{mathrsfs}%
\usepackage[title]{appendix}%
\usepackage{xcolor}%
\usepackage{textcomp}%
\usepackage{manyfoot}%
\usepackage{booktabs}%
\usepackage{algorithm}%
\usepackage{algorithmicx}%
\usepackage{algpseudocode}%
\usepackage{listings}%
\usepackage{enumitem}
\usepackage{float}
\usepackage{ragged2e}
\usepackage{parskip}
\theoremstyle{thmstyleone}%
\theoremstyle{thmstyletwo}%

\theoremstyle{thmstylethree}%

\begin{document}

% \title[Article Title]{A Multimodal Framework for Material Discovery: Leveraging Graphs and Large Language Models}

\title[Article Title]{MatterChat: A Multi-Modal LLM for Material Science}

%%=============================================================%%
%% GivenName	-> \fnm{Joergen W.}
%% Particle	-> \spfx{van der} -> surname prefix
%% FamilyName	-> \sur{Ploeg}
%% Suffix	-> \sfx{IV}
%% \author*[1,2]{\fnm{Joergen W.} \spfx{van der} \sur{Ploeg} 
%%  \sfx{IV}}\email{iauthor@gmail.com}
%%=============================================================%%

\author*[1]{\fnm{Yingheng} \sur{Tang}}\email{ytang4@lbl.gov}\equalcont{These authors contribute equally}

\author*[2]{\fnm{Wenbin} \sur{Xu}}\email{wenbinxu@lbl.gov}\equalcont{These authors contribute equally}

\author[3]{\fnm{Jie} \sur{Cao}}

\author*[4]{\fnm{Weilu} \sur{Gao}}\email{weilu.gao@utah.edu}

\author[2]{\fnm{Steven} \sur{Farrell}}

\author[5,6]{\fnm{Benjamin} \sur{Erichson}}

\author[5,6,7]{\fnm{Michael W.} \sur{Mahoney}}

\author[1]{\fnm{Andy} \sur{Nonaka}}

\author*[1]{\fnm{Zhi} \sur{Yao}}\email{jackie\_zhiyao@lbl.gov}

\affil[1]{Applied Mathematics and Computational Research Division, Lawrence Berkeley National Laboratory, Berkeley, CA, USA}

\affil[2]{National Energy Research Scientific Computing Center, Lawrence Berkeley National Laboratory, Berkeley, CA, USA}

\affil[3]{NSF National AI Institute for Student-AI Teaming, University of Colorado at Boulder, Boulder, USA}

\affil[4]{Department of Electrical and Computer Engineering, The University of Utah, Salt Lake City, UT, USA}

\affil[5]{Scientific Data Division, Lawrence Berkeley National Laboratory, Berkeley, CA, USA}

\affil[6]{International Computer Science Institute, Berkeley, CA, USA}

\affil[7]{Department of Statistics, University of California at Berkeley, Berkeley, CA, USA}
%% Sample for unstructured abstract %%
%%==================================%%

\abstract{Understanding and predicting the properties of inorganic materials is crucial for accelerating advancements in materials science and driving applications in energy, electronics, and beyond. Integrating material structure data with language-based information through multi-modal large language models (LLMs) offers great potential to support these efforts by enhancing human–AI interaction. However, a key challenge lies in integrating atomic structures at full resolution into LLMs. In this work, we introduce MatterChat, a versatile structure-aware multi-modal LLM that unifies material structural data and textual inputs into a single cohesive model. 
MatterChat employs a bridging module to effectively align a pretrained universal machine learning interatomic potential with a pretrained LLM, reducing training costs and enhancing flexibility.
Our results demonstrate that MatterChat significantly improves performance in material property prediction and human-AI interaction, surpassing general-purpose LLMs such as GPT-4. We also demonstrate its usefulness in applications such as more advanced scientific reasoning and step-by-step material synthesis.

% MatterChat creates a new structure-aware mutimodal LLM approach in general materials science applications. 
}


%%\pacs[JEL Classification]{D8, H51}

%%\pacs[MSC Classification]{35A01, 65L10, 65L12, 65L20, 65L70}

\maketitle

\section{Introduction}\label{sec1}

In-silico material discovery and design have traditionally relied on high-fidelity first-principles methods such as density functional theory (DFT)~\cite{doi:10.1021/jp960669l} and ab-initio molecular dynamics (AIMD)~\cite{marx2009ab} to accurately model atomic interactions and predict material properties. Despite their effectiveness, these methods face significant challenges due to their prohibitive computational cost, limiting their scalability for high-throughput screening across vast chemical spaces and for simulations over large length and time scales.
Moreover, many advanced materials remain beyond the reach of widespread predictive theories due to a fundamental lack of mechanistic understanding. These challenges stem from the inherent complexity of their chemical composition, phase stability, and the intricate interplay of multiple order parameters, compounded by the lack of self-consistent integration between theoretical models and multi-modal experimental findings.
As a result, breakthroughs in functional materials, such as new classes of correlated oxides, nitrides, and low-dimensional quantum materials, have largely been serendipitous or guided by phenomenological intuition rather than systematic, theory-driven design. 
Attempts to predict new materials and functionalities have often led to mixed results, with theoretically proposed systems failing to exhibit the desired properties when synthesized and tested. 
Achieving reliable, scalable, and predictive design of materials requires a paradigm shift.

With the rise of artificial intelligence (AI)  in materials science, there has been a surge of methods aiming to overcome these limitations, ranging from machine learning (ML) surrogate models~\cite{xie2018crystal, xu2022predicting}, ML interatomic potentials (MLIPs)~\cite{batzner20223, gasteiger2021gemnet, batatia2023foundation}, and generative models~\cite{2312.03687, xie2021crystal}. These models enable rapid predictions, accelerate large-scale simulations, and facilitate the generation of novel materials. As a result, they have significantly advanced fields such as energy storage~\cite{Ling2022}, electronics~\cite{Liu2022}, catalysis~\cite{doi:10.1021/acsomega.9b03673}, and biomedical applications~\cite{Sen2024}. Among these promising ML approaches, graph-based models in material science have become increasingly popular due to their versatile graph representation of atomistic systems, where each atom is represented as a node, and chemical bonds to neighboring atoms are represented as edges. Although these graph-based methods have shown success in accurately predicting material properties, they typically lack the capacity to handle tasks that require understanding scientific context, literature-based insights, and domain-specific language~\cite{Birhane2023}. In particular, these models do not support human–AI interaction through user prompts or textual descriptions, making it difficult to incorporate expert domain knowledge and user-specified requests to close the feedback loop.

This bottleneck has inspired a wave of exploration into how natural language processing (NLP), and particularly large language models (LLMs), might be leveraged to fill the gap.
LLMs like BERT~\cite{1810.04805}, GPT~\cite{2005.14165}, and newer open-source LLMs such as Mistral~\cite{2310.06825}, Llama~\cite{2302.13971} and DeepSeek~\cite{2501.12948} have shown substantial promise across different domains by interpreting and generating language. Trained on extensive datasets, these models can support some scientific tasks that require interpretive language capabilities, such as question-answering (QA)~\cite{2209.09513} and retrieving information from unstructured text sources~\cite{2212.05238}. 
In recent years, there has been several efforts incorporating LLM to solve material related problems~\cite{doi:10.1021/jacs.4c05840, 2409.02231}, either by leveraging pretrained LLMs or multi-modal LLMs. Unfortunately, these approaches rely on text-based representations, such as chemical formulas~\cite{doi:10.1021/jacs.4c05840}, Simplified Molecular Input Line Entry System (SMILES) strings~\cite{weininger1988smiles, 2409.02231}, textual descriptions~\cite{ock2023catalyst}, or Crystallographic Information File (CIF)~\cite{antunes2024crystal}, which lose the full resolution of atomic structures. Consequently, they exhibit inferior performance compared to pure graph-based models for predicting material properties~\cite{ock2023catalyst}, and they also potentially hinder other downstream tasks where structural information is crucial. Thanks to the steady development of MLIPs~\cite{batzner20223, gasteiger2021gemnet, batatia2023foundation}, particularly in their universal form (uMLIPs)~\cite{batatia2023foundation}, these models can now serve as atomistic pre-trained models capable of supporting a wide range of applications. The locality assumption underlying uMLIPs ensures that they effectively represent the local environment of each atom. Therefore, it is feasible to extract structural information from atom embeddings in a pretrained~uMLIP.

% A significant challenge remains: while LLMs excel at language tasks, they lack the ability to process material-specific structural information and align with specialized embeddings used for material properties. This limits their effectiveness in tasks requiring an integrated understanding of material compositions, properties, and insights from scientific literature. To address this, a hybrid framework is needed to bridge numerical representations of material structures with domain-specific knowledge in scientific text. Such a solution would enhance property predictions, support materials science QA tasks, and connect structural data with scientific knowledge for deeper insights.

%until here,  mention MatterChat here. is one model for many tasks emphaized?
In this work, we present MatterChat, a multi-modal large language model designed for materials science. 
MatterChat integrates material structure data with textual user queries, combining insights from materials science and NLP. It enhances the capabilities of large language models by overcoming their traditional limitations in quantitative predictions and improving the handling of scientific material-related tasks, as demonstrated through comparisons with other LLMs. 
MatterChat also maintains robust human–AI interaction capabilities, offering an intuitive interface for complex queries, compared to physical ML models. Furthermore, by leveraging deep, embedded knowledge from state-of-the-art pretrained LLMs, MatterChat enables advanced scientific reasoning and synthesis process guidance. 
The embedding visualization analysis indicates that MatterChat effectively preserves structure and property information. This has guided the adoption of a multi-modal Retrieval-Augmented Generation (RAG) approach that can enhance MatterChat’s robustness during material task inference.

\section{Results}\label{sec2}
\subsection{Overview of MatterChat}
\begin{figure}[!b]
    \centering
    \noindent\includegraphics[width=\textwidth]{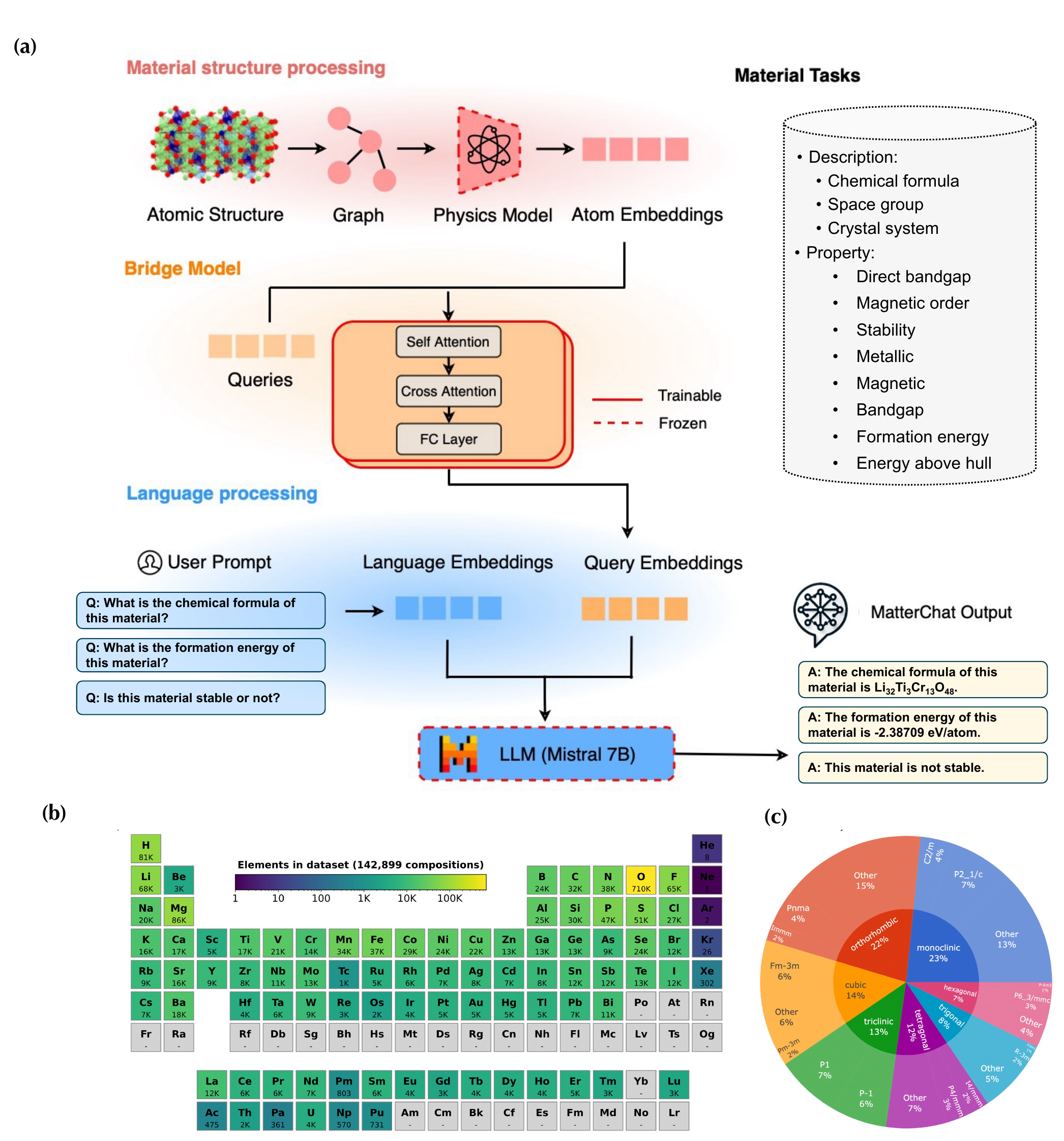}
    \caption{\textbf{Overview of MatterChat: a modular multi-modal LLM for material-based question-answering.} 
    (a) MatterChat architecture: The system includes a material encoder that generates atom embeddings and a LLM that processes language data. These components are connected by a trainable bridge model, which aligns material structure with natural language to support tasks such as material description and property prediction. (b) Elemental distribution across 142,899 compositions, representing the dataset’s compositional diversity. (c) Dataset distribution shown by space groups (outer ring) and crystal Systems (inner ring), illustrating structural variation within the dataset.}
    \label{fig:1}
\end{figure}

Figure \ref{fig:1}(a) presents the architecture of MatterChat, designed to process both material structures and user requests as inputs to generate text-based outputs for tasks such as material property prediction, structural analysis, and descriptive language generation. MatterChat consists of three core components: the Material Processing Branch, the Language Processing Branch, and the Bridge Model. The Material Processing Branch extracts atomic-level embeddings from material structures represented as graphs. These embeddings are then processed by the Bridge Model, which employs trainable queries to produce language model-compatible embeddings. Finally, the Language Processing Branch processes the user’s text-based prompt (e.g., “What is the formation energy of the material?”) into language embeddings. These embeddings are then combined with the query embeddings generated by the Bridge Model and fed into the LLM to produce the final output in text format. Below, we provide the details of each component.

\textbf{Material Processing Branch.}
The Material Processing Branch encodes material structures as graphs that capture the atomic local environment. We use CHGNet~\cite{Deng2023}, a state-of-the-art graph-based uMLIP model designed for crystal structures, to process these graphs. CHGNet is pretrained on a diverse dataset of materials, encompassing a wide range of symmetries, compositions, and bonding types, enabling it to effectively model complex atomic interactions and structural details. By capturing essential compositional features, such as atomic types and chemical bonds, along with spatial features like bond angles, CHGNet generates high-quality atom embeddings that are both physically meaningful and well-suited for downstream~tasks.

\textbf{Language Processing Branch.}
The Language Processing Branch is used to process user's text-based prompts, such as requests for property predictions, chemical formulas, space group information, or other material characteristics. We use the Mistral 7B LLM~\cite{2310.06825}, one of the latest open-source LLMs, chosen for its exceptional performance across a wide range of scientific and non-scientific tasks. This branch processes each prompt, transforming it into dense embeddings that capture the semantic content of the inquiry. These embeddings are then combined with the query embeddings processed by the bridge model using a structured fusion approach, allowing the model to effectively incorporate both textual and material information. This integration enables the LLM to generate precise and contextually relevant responses tailored to the user’s specific material-related prompts.

\textbf{Bridge Model.} To facilitate the integration between atom embeddings and the Language Processing Branch, we developed a bridge model inspired by the BLIP2 architecture~\cite{2301.12597} based on a multi-layer transformer framework. This bridge model includes 32 trainable query vectors that interact with atom embeddings using an alternating attention mechanism. Cross-attention in even-numbered layers extracts key features from the atom embeddings, while self-attention in odd-numbered layers enhances representational depth. This approach refines the atom embeddings into query embeddings that are most connected to text, as shown in Figure \ref{fig:1}a. Finally, these refined representations are mapped to LLM-compatible embeddings via a linear projection layer.

% \textbf{Fusion and Output:}
% After feature extraction, the outputs from the material structure and language branches are concatenated to form a unified multimodal embedding. This joint representation is then processed by subsequent layers of the LLM, generating rich, contextually informed responses that integrate both material structure and language inputs. The combined strengths of the pretrained graph foundation model, CHGNet, for atomic interaction modeling and the pretrained LLM, Mistral, for language processing, enable the model to produce highly relevant, domain-specific outputs tailored to material science queries.

Figures \ref{fig:1}(b)-(c) provide an overview of the dataset of crystalline structures used in our training set. Figure \ref{fig:1}(b) visualizes the material distribution on the periodic table, highlighting that the dataset evenly spans a diverse range of elements up to Plutonium. Figure \ref{fig:1}(c) depicts the distribution of crystalline structures by space group across the dataset.
The dataset was curated from the Materials Project~\cite{Jain2013} and contains 142,899 material structures. For each structure, we generated a corresponding text-based dataset encompassing 12 tasks: 3 descriptive tasks (chemical formula, space group, and crystal system) and 9 property prediction tasks. These property prediction tasks include metallicity, direct bandgap, stability, experimental observation (Exp Ob), magnetic status (Is Magnetic), magnetic order, formation energy, energy above the hull, and bandgap, as summarized in Figure \ref{fig:1}(a).
Further details regarding the training scheme, hyper-parameters, and dataset curation are provided in the Methods section.

\subsection{Multi-modal material task interactions with MatterChat}
\begin{figure}[!t]
    \centering
    \includegraphics[width=\textwidth]{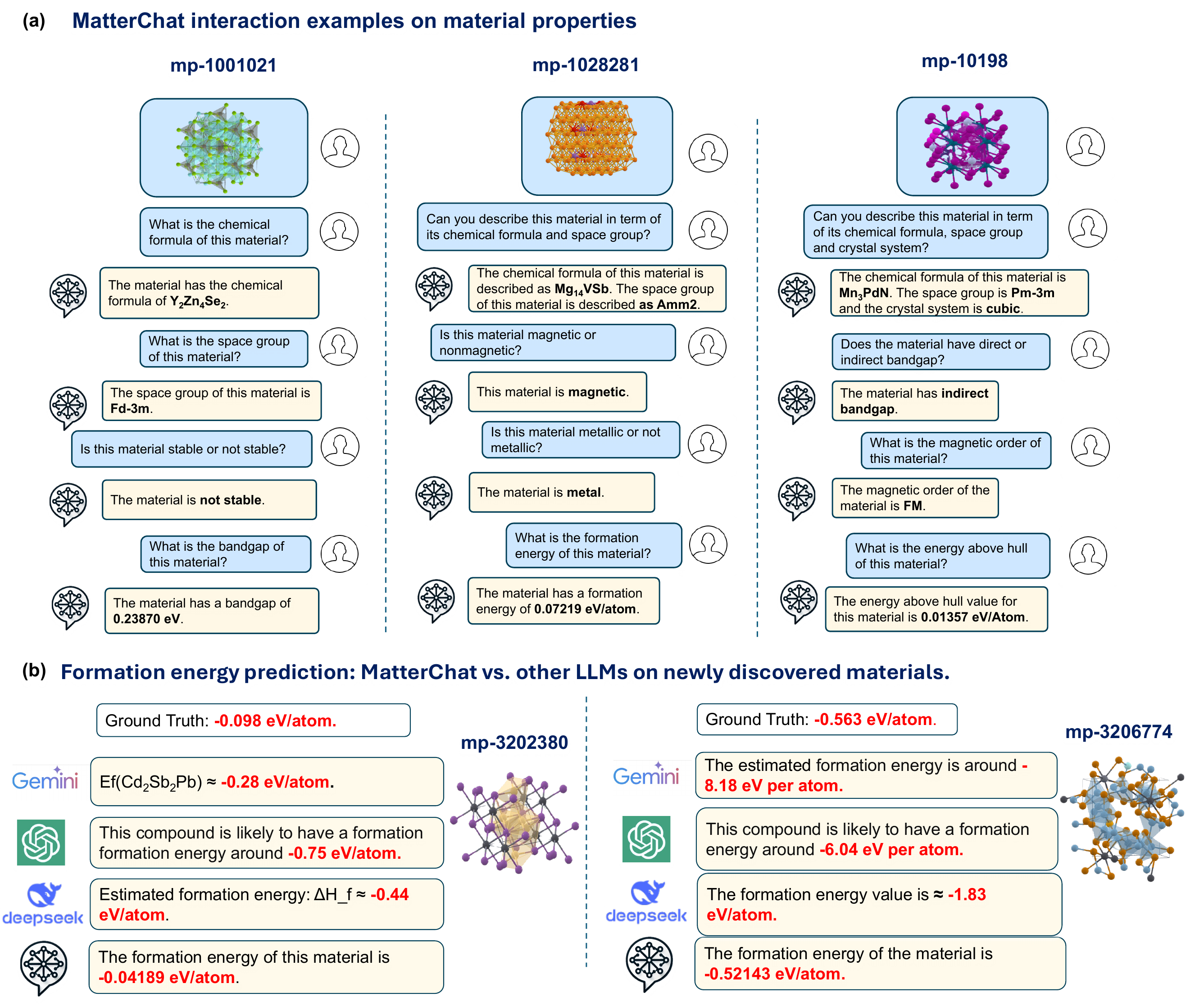}
    \caption{\textbf{MatterChat accurately predicts material properties and outperforms state-of-the-art LLMs.} (a) Illustration of multi-modal material property queries using MatterChat. The model accurately interprets user prompts to predict chemical formulas, crystallographic properties, stability, electronic bandgap, magnetic order, and energy metrics of materials. The three panels demonstrate the framework’s ability to address diverse material science inquiries, showing its alignment of graph-based and textual embeddings for precise question answering. (b) Comparative evaluation of formation energy predictions for newly discovered material from GNoME~\cite{Merchant2023}. Predictions from Matterchat compared against the ground truth values along with evaluations from commercial LLMs (Gemini~\cite{2312.11805}, GPT-4o~\cite{2410.21276} and DeepSeek~\cite{2501.12948}). The results show the accuracy and stability of the Matterchat in quantitative material evaluation tasks, which closely aligns with the ground truth, demonstrating its ability to integrate material graph embeddings for precise property prediction.}
    \label{fig:2}
\end{figure}
Figure \ref{fig:2} illustrates examples of human-AI interaction with MatterChat across a diverse range of material property prediction and analysis tasks. It shows MatterChat’s ability to effectively address a broad spectrum of user prompts ranging from fundamental material attributes (e.g., chemical formulas, space groups and crystal system) to complex material properties (e.g., thermal stability, bandgaps, formation energies and energy above the hull).
Figure \ref{fig:2}(a) shows three interactive examples of material property prompts from randomly selected materials from the Material Project Dataset.
The top left panel presents a human-AI query interface with MatterChat for the material with mp-id of ``mp-1001021''. It provides a detailed profile including the chemical formula $\rm{Y_2Zn_4Se_2}$, its crystalline structure denoted by the space group Fd-3m, and electronic properties such as a bandgap of 0.21350 eV. The interface also addresses the material’s lack of thermal stability.
The top mid panel shows the interaction example with the material with mp-id of ``mp-1028281''. It provides a comprehensive breakdown of the material’s composition attributes, including its chemical formula ($\rm{Mg_{14}VSb}$) and its space group (Amm2). The interaction further predicts that the material is both magnetic and metallic, and its formation energy is estimated at 0.05912 eV/atom.
The top right panel provides an interaction example with MatterChat of the material with mp-id of ``mp-10198''. This panel informs the user's query about the chemical composition $\rm{Mn_3PdN}$ and its cubic crystal structure, with the space group classified as Pm-3m. Additionally, it estimated that the material possesses an indirect bandgap, which is an important characteristic for applications in electronics. MatterChat also accurately predicts the ferromagnetic (FM) magnetic behaviors that the material exhibits, and it mentions its energy above hull value at 0.03571 eV/atom.
In the bottom panel, we present a comparative evaluation of MatterChat’s performance on formation energy evaluation tasks for newly discovered materials from GNoME~\cite{Merchant2023}. The model was compared against commercial LLMs, Gemini~\cite{2312.11805}, GPT-4o~\cite{2410.21276} and DeepSeek~\cite{2501.12948}. 
The results show MatterChat’s superior accuracy in estimating formation energies, consistently delivering predictions closer to the ground truths. For example, MatterChat’s formation energy predictions for ``mp-3202380'' and ``mp-3206774'' show a remarkable alignment with the ground truth values.
These results demonstrate MatterChat’s ability to integrate structural and textual data seamlessly for a wide range of material property tasks.

\begin{figure}[!t]
    \centering
    \includegraphics[width=\textwidth]{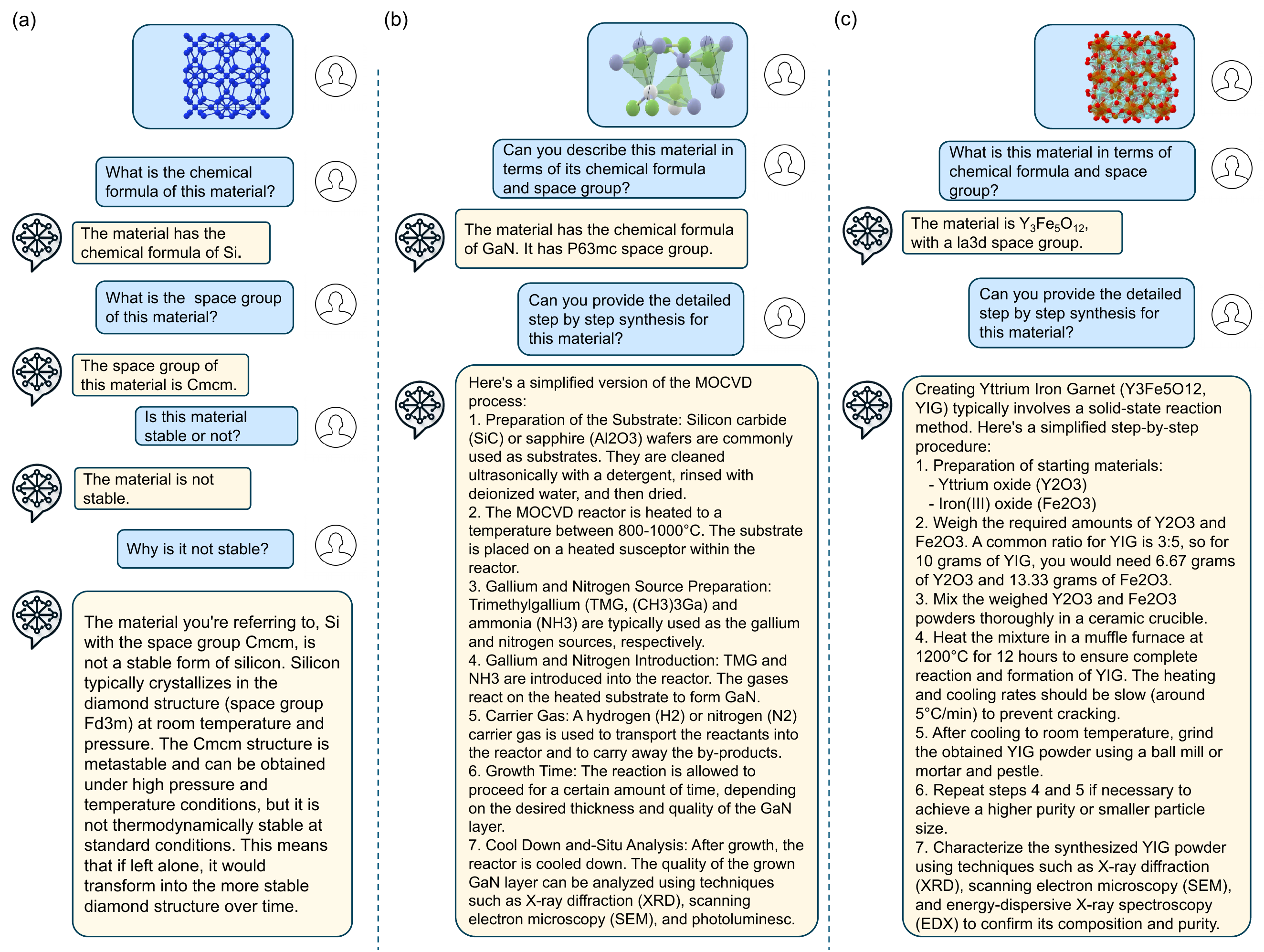}
    \caption{\textbf{MatterChat has the ability to solve more sophisticated tasks inherited from the pretrained LLM.} (a) Material property query for silicon (Si), including its chemical formula, space group, stability, and the reasoning for why it is not stable under standard conditions. (b) Highlights a material query for Gallium Nitride (GaN), providing its chemical formula, space group, and a step-by-step synthesis procedure using methods like HVPE, MOCVD, and MBE. (c) Material query interaction, Yttrium Iron Garnet (YIG, $\rm {Y_3Fe_5O_{12}}$), detailing its chemical formula, space group, and a simplified step-by-step synthesis procedure using the solid-state reaction method.}
    \label{fig:3}
\end{figure}

Figure \ref{fig:3} demonstrates MatterChat’s advanced reasoning capabilities, showing how it leverages the comprehensive knowledge base of LLMs to address complex materials science challenges. By using a multi-modal query system, MatterChat effectively combines material structure data with textual reasoning. This integration facilitates a working memory scheme~\cite{2408.13654}, which enables the model to provide domain-specific reasoning, detailed synthesis procedures, and explanations that are deeply grounded in the structural properties of materials. 
Panel (a) presents the chat log for silicon with the space group of cmcm. MatterChat not only retrieves the chemical formula and correct space group, but it also provides a rationale for the structural instability of this silicon phase. The model explains that the cmcm space group exhibits higher energy per unit cell compared to the thermodynamically stable cubic diamond structure of silicon, making it less likely to occur under standard conditions.
Panel (b) illustrates an interaction regarding a popular semiconductor material Gallium Nitride (GaN). Here, MatterChat accurately identifies the chemical formula and space group (P63mc) and constructs a detailed synthesis procedure that aligns with established methods such as molecular beam epitaxy (MBE), metal-organic chemical vapor deposition (MOCVD), and hydride vapor phase epitaxy (HVPE), demonstrating the model’s ability to apply structural and contextual knowledge in generating practical scientific outputs.
Figure \ref{fig:3}(c) explores an interaction for a widely used ferrite material, Yttrium Iron Garnet (YIG, $\rm{Y_3Fe_5O_{12}}$). MatterChat is able to take the structure and generate detailed text descriptions. Additionally, MatterChat can further generate the synthesis protocol from the preparation of initial materials, their precise mixing ratios, the required sintering conditions, and subsequent characterization techniques, such as X-ray diffraction (XRD) and scanning electron microscopy (SEM). This example  demonstrates MatterChat’s capability to synthesize and apply domain-specific knowledge effectively, aligning closely with established scientific practice~\cite{10.1063/1.4973199}.

\subsection{MatterChat extracted embeddings contain structural and property information}
\begin{figure}[!t]
    \centering
    \includegraphics[width=\textwidth]{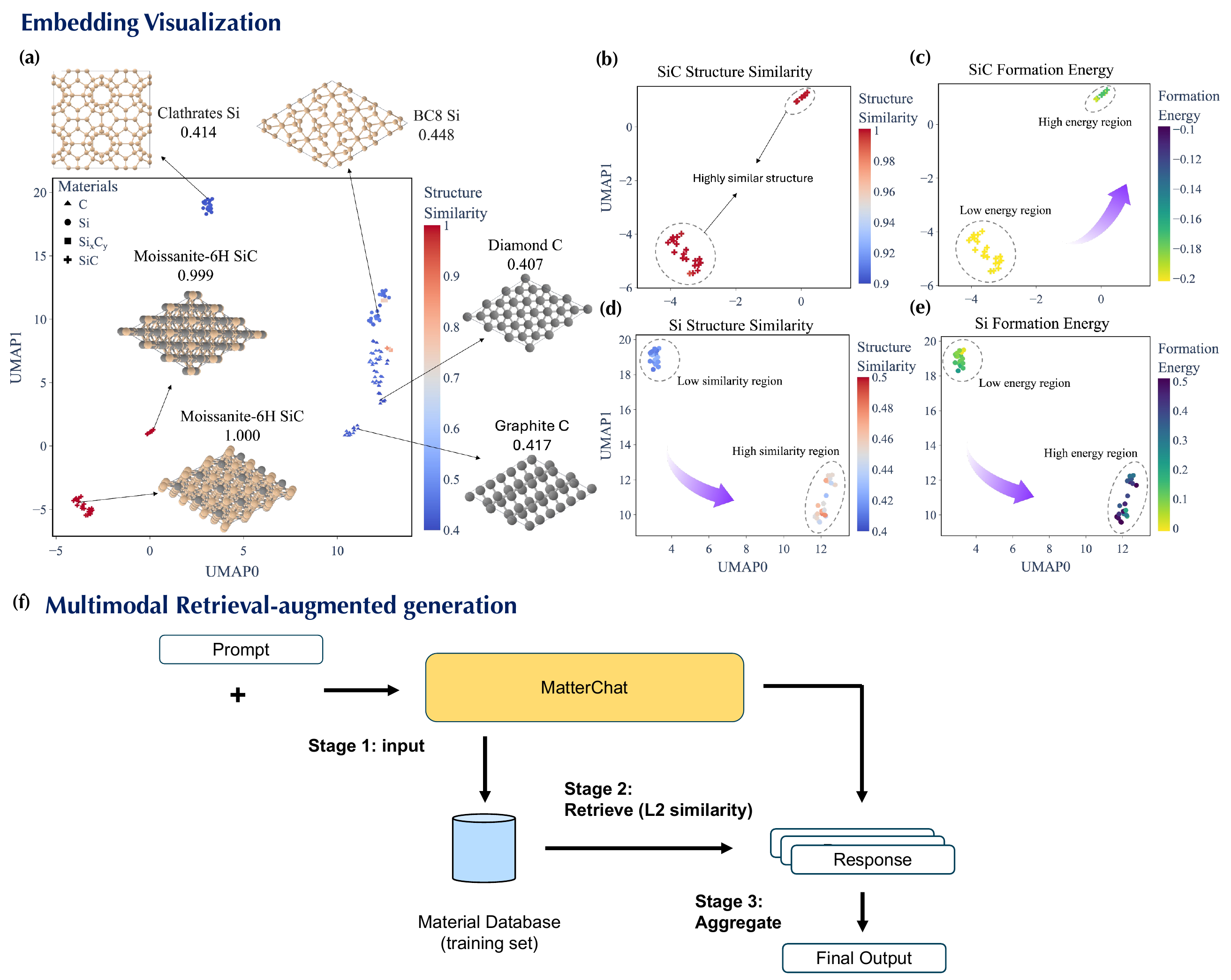}
    \caption{\textbf{UMAP visualization of structural embeddings extracted from the bridge model.} (a) Visualization of samples containing $\rm Si$ and $\rm C$ elements from the Material Project dataset, showing how materials cluster based on their structural embeddings extracted from the bridge model. The value indicates the structural similarity calculated using the SOAP descriptor in combination with the REMatch kernel (see Methods for further details). (b, c) Visualizations of SiC subgroup color-coded by structural similarity and formation energy. The two clusters exhibit high structural similarity, with formation energy further assisting in distinguishing between them. (d, e) Visualizations of Si subgroup color-coded by structural similarity and formation energy. Two clusters demonstrate a smooth transition in both structural similarity and formation energy, indicating that both factors captured by the structural embeddings contribute to the observed clustering. (f) Proposed Multi-modal Retrieval-Augmented Generation (RAG) for robust prediction.}
    \label{fig:4}
\end{figure}
We further explore MatterChat's ability to leverage material structural information by providing a detailed visualization/clustering analysis with the UAMP dimension reduction technique~\cite{1802.03426}. Figures \ref{fig:4}(a)-(e) show comprehensive visualizations of embeddings processed by the bridge model, with all material samples that contain silicon (Si), carbon (C), and their composites compounds (e.g., $\rm SiC$, $\rm Si_xC_y$) from the Material Project database~\cite{https://doi.org/10.6084/m9.figshare.23713842.v2}. UMAP was used to reduce the embeddings from an original 4096 dimensions to two dimensions, with the x and y axes corresponding to the first and second reduced dimensions, respectively.

Figure \ref{fig:4}(a) presents the visualizations containing all the selected materials; each sample is color-coded with a structure similarity score~\cite{HIMANEN2020106949}. The clustering generally follows distinctions in chemical compositions. Additionally, materials with the same atomic composition are grouped into separate clusters based on crystalline structural differences (e.g., Carbon with Diamond vs. Graphite crystalline structure).
Figures \ref{fig:4}(b) and \ref{fig:4}(d) show zoomed-in visualizations of clustering results for materials consisting exclusively of Si and SiC compositions. Figure \ref{fig:4}(d) shows the gradient of structure similarity scores, ranging from blue (low similarity) to red (high similarity), demonstrating how closely related structural features result in spatial proximity within the embedding space. However, an interesting exception is observed with SiC (see Figure \ref{fig:4}(b)): despite its identical composition and similar structural phases, two distinct clusters of SiC emerge, suggesting that factors beyond composition and structure alone influence their separation.
To further explore factors that influence clustering, we labeled the samples according to their formation energy, with results displayed in Figures \ref{fig:4}(c) for SiC and \ref{fig:4}(e) for Si.
These figures clearly show a trend from low to high formation energy. This analysis reveals that clusters grouped by structural similarity also align closely in terms of formation energy. Such findings indicate that the model’s ability to produce embeddings that not only differentiate structural characteristics but also correlate with key material properties.
Given that the embeddings derived from bridge model preserve both material structure and property-relevant information, we implemented a multi-modal Retrieval-Augmented Generation (RAG) mechanism during inference, as illustrated in Figure \ref{fig:4}(f). Instead of relying solely on a single output from MatterChat for each query-sample pair, we now retrieve additional information of two more samples from the material pool (training set). This retrieval is based on the L2 similarity between the embeddings of the sample material and those in the pool. After that, we aggregate all three results to get the final output by applying a majority-voting strategy for classification tasks and averaging for quantitative tasks. Such a method could further enhance the overall robustness of MatterChat across different tasks. The details of the visualization method are provided in the Method Section.
% By incorporating insights from the two most structurally similar materials in the training set, we enhance the prediction process. For classification tasks, we apply a majority-voting strategy for the final output; for regression tasks, we average the predictions for improved accuracy.

\subsection{Comprehensive quantitative analysis for all material tasks}
To provide a more thorough evaluation of MatterChat, we conducted a comprehensive analysis of its performance across nine material property tasks on the evaluation set (14290 samples). This analysis compares MatterChat with two open-source large language models (LLMs), Vicuna~\cite{vicuna2023} and Mistral~\cite{2310.06825}, as well as two physical ML models, SchNet~\cite{1706.08566} and CHGNet~\cite{Deng2023}.

\begin{figure}[!t]
    \centering
    \includegraphics[width=\textwidth]{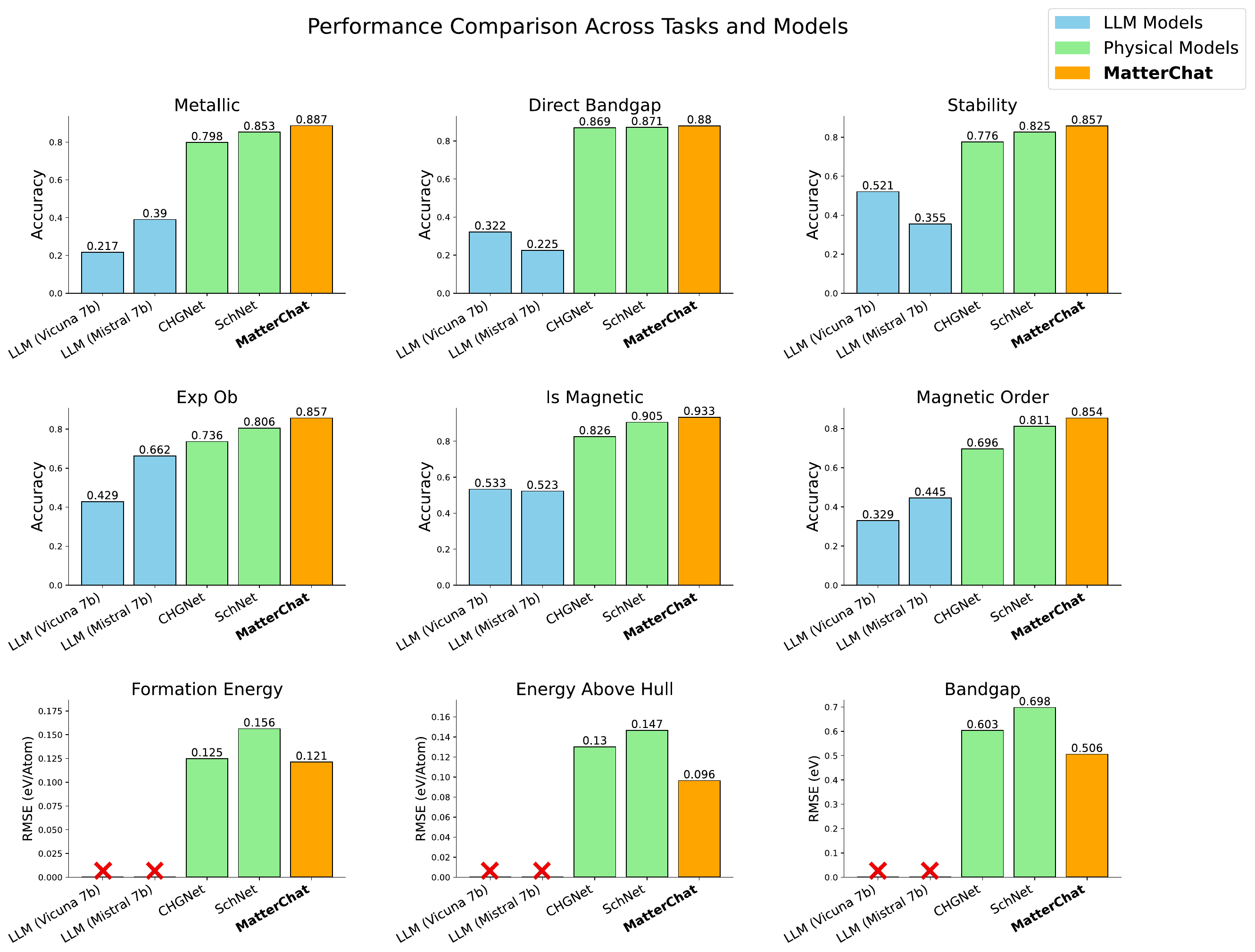}
    \caption{\textbf{Performance comparison of MatterChat, open-source LLMs (Vicuna, Mistral), and physical pre-trained models (SchNet, CHGNet) across nine material property tasks.} (a)–(f) show classification task accuracies, where MatterChat consistently outperforms other models. Panels (g)–(i) present root mean absolute error (RMSE) results for numerical property predictions, demonstrating MatterChat’s superior precision in formation energy, energy above the hull, and bandgap tasks.}
    \label{fig:5}
\end{figure}

\textbf{Classification Tasks (Figure \ref{fig:5}(a)–(f))}
Six out of the nine material property tasks are classification tasks, evaluated using prediction accuracy as the metric. These tasks include metallicity, bandgap type, stability, magnetic properties, and experimental observables. As shown in the top two rows of Figure \ref{fig:5}, MatterChat demonstrates improvements over all tested models. Compared to state-of-the-art LLMs, MatterChat achieves substantially higher accuracy, benefiting from its ability to integrate graph-based material structure data with natural language reasoning. Moreover, MatterChat even outperforms physical ML models such as SchNet and CHGNet, the latter being the built-in model used in the Material Processing Branch, showing its superior capability in handling multi-modal inputs for material classification.\\
\textbf{Numerical Property Prediction (Figure \ref{fig:5}(g)–(i))}
The remaining three tasks—formation energy, energy above the hull, and bandgap prediction—are numerical property prediction tasks. For these tasks, the root mean squared error (RMSE) between predicted values and ground truth is used as the evaluation metric. Due to the inherent limitations of LLMs in producing accurate numerical predictions~\cite{2410.13857}, their performance is not included in the bar plots, as the errors were excessively large. Instead, Figure \ref{fig:5} compares MatterChat with the physical ML models.
The results show that MatterChat achieves an overall lower RMSE error in the numerical property tasks, outperforming both SchNet and CHGNet. This demonstrates MatterChat’s ability to handle highly quantitative material property predictions with precision, further validating the effectiveness of its multi-modal integration approach. 
Such analysis demonstrates MatterChat’s versatility and robustness across diverse material property tasks. By outperforming both state-of-the-art LLMs and physical ML models, MatterChat establishes itself as a powerful tool for materials science applications, offering accurate predictions for both classification and numerical property tasks. These results demonstrate the framework’s potential to accelerate material discovery and deepen our understanding of material properties.

\subsection{Comparative Study}
We conducted a comprehensive comparative study to evaluate the effectiveness of our multi-modal LLM approach against several alternative methods. The summary of inference results on the test set across all material property prediction tasks is presented in Table~\ref{table:1}.\\
The first column shows results from training the multi-modal LLM using a Simple Adapter with Low-Rank Adaptation (LoRA) finetuning\cite{hu2022lora}, where lightweight adapter layers and the LLM (Mistral-7B) are updated during training, following the method in~\cite{2304.08485}.
The second column presents a pure LLM baseline, in which the LLM is finetuned with LoRA on serialized CIF content and user queries treated as tokenized text prompts—without any structural encoding or auxiliary modality.
The third column corresponds to our proposed method based on the Bootstrapping Approach~\cite{2301.12597}, where only the bridge module is trained, while both the graph encoder and the LLM remain frozen. This strategy focuses on aligning material structures with language representations efficiently, avoiding the need for extensive finetuning of the large-scale language model.
All methods were trained on the same dataset under identical conditions and evaluated on the same test splits. The results demonstrate that our multi-modal bootstrapping approach achieves superior performance compared to both baselines. Notably, it does so without modifying the pretrained components, highlighting its efficiency and practicality for structure-aware property prediction tasks using LLM.
We also include results from our approach augmented with multi-modal retrieval-augmented generation (RAG) during inference. This variant further slightly improves inference performance by having taking more similar material samples into consideration during the inference.

\begin{table}[!t]
\centering
\caption{Comparison of material property prediction performance across different multi-modal frameworks and RAG-enhanced inference}
\label{tab:combined}
\begin{tabular}{@{}lcccc@{}}
\toprule
\textbf{Task} & \textbf{Simple Adapter w LoRA} & \textbf{LoRA LLM only} & \textbf{MatterChat} & \textbf{MatterChat w/ RAG} \\
\midrule
Metallic (Accuracy) & 0.6373 & 0.6864 & 0.8683 & \textbf{0.8873} \\
Direct Bandgap (Accuracy) & 0.8629 & 0.7839 & 0.8753 & \textbf{0.8797} \\
Stability (Accuracy) & 0.7418 & 0.7944 & 0.8515 & \textbf{0.8573} \\
Exp Ob (Accuracy) & 0.7171 & 0.6549 & 0.8504 & \textbf{0.8570} \\
Is Magnetic (Accuracy) & 0.8339 & 0.6833 & \textbf{0.9368} & 0.9333 \\
Magnetic Order (Accuracy) & 0.7759 & 0.4238 & \textbf{0.8570} & 0.8535 \\
Formation Energy (RMSE) & 0.4105 & 1.8059 & 0.1500 & \textbf{0.1212} \\
Energy Above Hull (RMSE) & 0.4415 & 0.4051 & 0.1053 & \textbf{0.0964} \\
Bandgap (RMSE) & 1.2516 & 1.4725 & 0.5590 & \textbf{0.5058} \\
\botrule
\end{tabular}
\begin{tablenotes}
\small
\item \textbf{Note:} RMSE units — Formation Energy and Energy Above Hull in eV/Atom, Bandgap in eV.
\end{tablenotes}
\label{table:1}
\end{table}
% Table \ref{table:2} is another comaprtive study where we evaluate the performance improvement for using the multi-modal RAG during inference.

% \begin{table}[!t]
% \centering
% \caption{Material property prediction task results with RAG}\label{tab:reordered}
% \begin{tabular}{@{}lcc@{}}
% \toprule
% \textbf{Task} & \textbf{MatterChat} & \textbf{MatterChat w RAG} \\
% \midrule
% Metallic(Accuracy) & 0.8683 & \textbf{0.878} \\
% Direct Bandgap(Accuracy) & 0.8753 & \textbf{0.872} \\
% Stability(Accuracy) & 0.8369 & \textbf{0.855} \\
% Exp Ob(Accuracy) & 0.8119 & \textbf{0.857} \\
% Is Magnetic(Accuracy) & \textbf{0.9343} & 0.9322 \\
% Magnetic Order(Accuracy) & 0.8477 & \textbf{0.8507} \\
% Formation Energy(RMSE) & 0.1526 & \textbf{0.1225} \\
% Energy Above Hull(RMSE) & 0.1095 & \textbf{0.0949} \\
% Bandgap(RMSE) & 0.5428 & \textbf{0.5070} \\
% \botrule
% \end{tabular}
% \label{table:2}
% \end{table}

\section{Discussion}\label{sec4}
In this study, we present MatterChat, a multi-modal framework that achieves superior performance in material properties prediction and scientific reasoning tasks by leveraging a more effective representation of materials. 
A key innovation of MatterChat is its ability to leverage existing advancements in both material science and language modeling by integrating a pretrained material foundation encoder with a pretrained large language model (LLM). Rather than training an entire model from scratch, MatterChat achieves strong performance by training only a lightweight bridge model, efficiently aligning material structure representations with textual understanding while maintaining high accuracy across diverse material science tasks.
Moreover, MatterChat is designed for multitask learning, enabling it to handle both classification and numerical property prediction. This capability allows the framework to tackle a diverse range of material science tasks within a unified model.
Another advantage of our approach is the use of graph-based structural embeddings instead of relying solely on Crystallographic Information File (.cif) text input. While CIF files encode atomic structures, their text-based format relies entirely on attention mechanisms, which can struggle to explicitly capture geometric symmetries and increase computational overhead due to lengthy tokenization. By processing atomic graphs directly, MatterChat effectively preserves material symmetry and spatial relationships, leading to more accurate structure-property learning while maintaining computational efficiency.
% In this study, we present MatterChat, a multi-modal framework that demonstrates superior performance in material property prediction and scientific reasoning tasks compared to traditional models. This performance can be attributed to its ability to integrate structural material data with language-based information, creating a more comprehensive representation of materials. Traditional approaches often rely on text-based representations, such as Crystallographic Information Files (.cif), to encode atomic structures. While .cif files effectively capture detailed structural information, their static format may limit the model’s capacity to learn complex structure-property relationships when used directly as input. In contrast, MatterChat’s multi-modal approach processes structural embeddings derived from atomic graphs alongside textual data, enabling dynamic interactions between atomic configurations and contextual scientific knowledge. This integration enhances the model’s generalization ability across diverse material tasks, as evidenced by its improved accuracy in classification and numerical property prediction~tasks. 

% \subsection*{Limitation}
While MatterChat demonstrates strong performance in material property prediction and analysis, several limitations present opportunities for improvement. The alignment between graph-based material embeddings and language representations remains behavior-driven rather than fully representation-level aligned, requiring further investigation. Additionally, the dataset’s fixed paraphrased queries may limit linguistic diversity, suggesting the need for more comprehensive textual descriptions and a broader range of material properties. Using a frozen LLM, primarily trained on general-purpose text, may restrict domain-specific reasoning and can potentially benefit from finetuning on material science datasets and literature. 
Future work could focus on three key directions to enhance MatterChat. First, improving representation-level alignment between modalities by incorporating contrastive loss during end-to-end instructive fine-tuning, in addition to pretraining, could further reinforce structural-textual consistency. Second, expanding the dataset to improve generalization by increasing linguistic diversity, as the current fixed set of paraphrased queries may limit variability. Incorporating more comprehensive descriptions and a broader range of material properties would enable the model to handle diverse material science tasks more effectively. Finally, to further enhance MatterChat’s generative capabilities, integrating a graph generative module on top of the existing model could enable novel material structure discovery, extending its utility from property prediction to full-scale material design.

\section{Method}\label{sec2}

\subsection{Dataset curation}\label{subsubsec2}
In this work, we curated a comprehensive dataset from the Materials Project Trajectory (MPtrj) dataset~\cite{https://doi.org/10.6084/m9.figshare.23713842.v2}, focusing specifically on relaxed samples. By selecting these stable configurations, rather than complete trajectory data, we ensure that the dataset captures the equilibrium states of materials, which are more relevant for downstream tasks such as material property prediction. The final dataset consists of 142,899 high-quality samples, offering a rich and diverse representation of inorganic materials.

To facilitate effective model training and evaluation, we partitioned the dataset into training and testing subsets using a 9:1 split ratio. This ensures that a substantial portion of the data is available for learning, while maintaining a dedicated portion for rigorous performance validation, allowing us to assess the generalization capabilities of the model.

In addition to the relaxed structural data, we retrieved detailed material property information using the Materials Project API~\cite{Jain2013}. Each material is retrieved by a unique mp-id and is enriched with a variety of key descriptors that span both structural and electronic properties. These include:

\begin{itemize}
    \item \textbf{Structure}: The full atomic structure of the material, detailing atomic positions and bonding.
    \item \textbf{Chemical formula}: The overall chemical composition.
    \item \textbf{Space group}: The crystallographic space group of the material, reflecting its symmetry properties.
    \item \textbf{Crystal system}: The broader classification of the material’s crystal structure.
    \item \textbf{Metallicity}: An indicator of whether the material is metallic or insulating.
    \item \textbf{Magnetic properties}: Including whether the material is magnetic and its magnetic ordering (e.g., ferromagnetic, antiferromagnetic).
    \item \textbf{Experimental observables}: Properties that can be compared directly with experimental data.
    \item \textbf{Direct bandgap}: The direct bandgap energy, a key property for semiconductors.
    \item \textbf{Stability}: Indicating whether the material is thermodynamically stable.
    \item \textbf{Energy above hull}: A measure of how stable the material is compared to other phases.
    \item \textbf{Bandgap}: The electronic bandgap, an important factor in determining a material’s electronic properties.
    \item \textbf{Formation energy}: The energy required to form the material from its constituent elements.
\end{itemize}

These attributes offer a comprehensive view of each material, encompassing both its structural arrangement and electronic behavior. By integrating this wealth of data, our model is capable of capturing complex material-property relationships, supporting tasks such as bandgap prediction, stability analysis, and metallicity determination. This dataset not only provides a robust foundation for training ML models, but it also contributes to broader efforts in materials discovery and property optimization.

\subsection{Training Detail}
MatterChat employs a bootstrapping strategy commonly used in multi-modal learning for vision-language tasks, adapted here for material science applications. The training process consists of two main stages: pretraining to align material structures with descriptive text; and finetuning for both descriptive and property prediction tasks with the LLM module integrated (see Figure S2 in the Supplementary Material).
The pretraining phase aims to establish a foundational alignment between material structures and descriptive text. In this stage, the model connects a frozen graph encoder with pairs of graph data and corresponding textual descriptions, without attaching the LLM module. Here, the bridge model acts as a text generator, learning to extract descriptive graph representations that effectively capture structural information relevant to the text data.

This stage consists of three core optimizing targets, each with distinct interaction mechanisms between graph embeddings and text, while maintaining a consistent input format:

\begin{enumerate}
    \item \textbf{Graph-Text Correlation Learning (Contrastive Loss).} \\
    This task aligns graph and text representations by maximizing the similarity between matched graph-text pairs and minimizing it for mismatched pairs. A contrastive loss is employed:
    \begin{equation}
        \mathcal{L}_{\text{correlation}} = - \sum_{i=1}^{N} \log \frac{\exp(\text{sim}(q_i, t_i)/\tau)}{\sum_{j=1}^{N} \exp(\text{sim}(q_i, t_j)/\tau)} ,
    \end{equation}
    where \( q_i \) and \( t_i \) represent the graph and text embeddings, respectively, and \( \tau \) is the temperature parameter controlling the distribution’s sharpness.

    \item \textbf{Graph-Driven Text Prediction (Conditional Language Modeling Loss).} \\
    The bridge model generates descriptive text based on graph data, conditioned through attention mechanisms. The loss function is defined as:
    \begin{equation}
        \mathcal{L}_{\text{prediction}} = - \sum_{t=1}^{T} \log P(y_t | y_{<t}, Q) ,
    \end{equation}
    where \( Q \) represents graph query features, and \( y_t \) is the token at position \( t \) in the output sequence.

    \item \textbf{Graph-Text Association (Binary Cross-Entropy Loss).} \\
    This task predicts whether each graph-text pair is correctly matched. A binary cross-entropy loss with hard negative sampling is applied:
    \begin{equation}
        \mathcal{L}_{\text{association}} = - \sum_{i=1}^{N} \left( y_i \log(s_i) + (1 - y_i) \log(1 - s_i) \right) ,
    \end{equation}
    where \( s_i \) is the model’s prediction score, and \( y_i \) indicates whether the pair is matched (1) or not (0).
\end{enumerate}

The total pretraining loss is the sum of the individual task losses:
\begin{equation}
    \mathcal{L}_{\text{total}} = \mathcal{L}_{\text{correlation}} + \mathcal{L}_{\text{prediction}} + \mathcal{L}_{\text{association}} .
\end{equation}

After pretraining, the model undergoes instructive finetuning to optimize its performance on both descriptive and property prediction tasks. In this stage, the pretrained bridge model is integrated with the LLM to enhance multi-modal learning. A fully connected layer is introduced between the bridge model’s output and the LLM’s input. The finetuning phase includes 12 multi-modal subtasks, including 3 material description tasks and 9 property prediction tasks. Description tasks refine the model’s ability to link structural features with detailed textual explanations, while property prediction tasks focus on improving quantitative accuracy in material property estimation. 
Finetuning is guided by a supervised cross-entropy loss defined as:
\begin{equation}
    \mathcal{L}_{\text{finetune}} = - \sum_{i=1}^{N} \sum_{j=1}^{T} y_{i,j} \log P(y_{i,j} | x_i) ,
\end{equation}
where \( y_{i,j} \) represents the ground truth token for the \( j \)-th position of the \( i \)-th sample, and \( P(y_{i,j} | x_i) \) is the model’s predicted probability of the correct token given the multi-modal input \( x_i \).

In the pretraining stage, the model is trained using the AdamW optimizer with a learning rate of \(2 \times 10^{-4}\), with cosine decay scheduler and linear warmup starting from \(1 \times 10^{-6}\). A weight decay of 0.05 is applied to regularize the model, with a batch size of 32 and gradient accumulation over 5 steps to manage computational efficiency. Mixed-precision training is enabled to improve performance and reduce memory usage. The model is trained for $\sim$ 25 epochs, with checkpoints saved every 2000 iterations. During the finetuning stage, the AdamW optimizer is again used with a learning rate of \(2 \times 10^{-4}\), featuring a warmup phase to \(1 \times 10^{-4}\) followed by decay to \(1 \times 10^{-5}\). The batch size is set to 8, with gradient accumulation over 16 batches to effectively increase the batch size. 
Finetuning runs for 50 epochs, with checkpoints saved every 300 steps and at the end of each epoch. Additionally, distributed training is implemented using 4 GPUs per node across 8 nodes, leveraging the Distributed Data Parallel (DDP) strategy to enhance training efficiency and scalability.

\subsection{Embedding Visualization}\label{Embedding Visualization}
The visualization leverages UMAP to reveal chemical insights encoded in the material embeddings that are extracted from the bridge model in a lower-dimensional space. To prepare the data, each high-dimensional embedding, originally structured as (32, 4096), is first flattened into a single vector, capturing the essential features of the material. UMAP is then applied to this set of vectors with number of components equals 2, reducing the data to two dimensions to enable visual interpretation, with random state is set to 1 to ensure consistency in the layout across runs.

Structural similarity scores are computed using the Smooth Overlap of Atomic Positions (SOAP) descriptor~\cite{bartok2013representing}, combined with the Regularized Entropy Match Kernel (REMatch)~\cite{de2016comparing, musil2018machine} to capture the structural characteristics within material embeddings. SOAP is a local atomic environment descriptor that encodes atomic geometries by expanding a Gaussian-smeared atomic density locally, using orthonormal functions derived from spherical harmonics and radial basis functions.  From local descriptors to structure matching, we use REMatch kernel on top of SOAP descriptor. The REMatch kernel considers the best matching of local environments and employs an averaging strategy to enhance structural comparison.  For SOAP construction, we consider periodic boundary conditions. The cutoff radius for the local region ($r_{cut}$), the number of radial basis functions ($n_{max}$), and the maximum degree of spherical harmonics ($l_{max}$) are set to 6 \r{A}, 8 \r{A}, and 6 \r{A}, respectively. 
For the REMatch kernel, the entropic penalty ($\alpha$) is set to 1, and the convergence threshold is set to $(1 \times 10^{-6})$. A linear pairwise metric is used for the local similarity calculation.

% \section{Conclusion and future work}\label{sec4}

\section*{Data availability}
Upon publication, all data that support the plots within this paper and other findings of this study will be available on a public \emph{GitHub} repository.  
% \michael{Will this be done with the arXiv submission.}
% \yingheng{When we first submit to arXiv, the code may not be ready. We’ll continue to refine and clean up the code during this period. Once everything is finalized, we will make both the data and the code publicly available in a repository to comply with the journal submission process, which requires code disclosure.}

\section*{Code availability}
Upon publication, all codes that support the plots within this paper and other findings of this study will be available on a public \emph{GitHub} repository. 
% \michael{Will this be done with the arXiv submission.}
% \yingheng{When we first submit to arXiv, the code may not be ready. We’ll continue to refine and clean up the code during this period. Once everything is finalized, we will make both the data and the code publicly available in a repository to comply with the journal submission process, which requires code disclosure.}
\section*{Acknowledgements}
Y.T., Z.Y., and A.N. were supported by Laboratory Directed Research and Development (LDRD) funding from Berkeley Lab, provided by the Director, Office of Science, of the U.S. Department of Energy under Contract No.\ DE-AC02-05CH11231.  This research used resources of the National Energy Research Scientific Computing Center (NERSC), a DOE Office of Science User Facility supported by the Office of Science of the U.S. Department of Energy under Contract No.\ DE-AC02-05CH11231 and under NERSC GenAI award under No.\ DDR-ERCAP0030541. W.G. acknowledges support from the National Science Foundation through Grant No. 2235276.

\section*{Author Contributions Statement}
Y.T.\ and W.G.\ conceived the idea, with W.X.\ contributing enhancements and Y.Z.\ providing additional support. Y.Z.\ supervised the project. Y.T.\ constructed overall ML framework and performed ML training/inference. W.X performed the physical model training and visualization. Y.T.\ and W.X.\ wrote the manuscript with the help of Y.Z., J.C., A.N, S.F., B.E. and M.M. 

\bibliographystyle{unsrt}
\bibliography{main}%
\end{document}

% \title[Article Title]{A Multimodal Framework for Material Discovery: Leveraging Graphs and Large Language Models}

\title[Article Title]{Supplementary Information:A Multimodal Framework for Material Discovery: Infusing Large Language Models with Atomic Structures}

%%=============================================================%%
%% GivenName	-> \fnm{Joergen W.}
%% Particle	-> \spfx{van der} -> surname prefix
%% FamilyName	-> \sur{Ploeg}
%% Suffix	-> \sfx{IV}
%% \author*[1,2]{\fnm{Joergen W.} \spfx{van der} \sur{Ploeg} 
%%  \sfx{IV}}\email{iauthor@gmail.com}
%%=============================================================%%

\author*[1]{\fnm{Yingheng} \sur{Tang}}\email{ytang4@lbl.gov}\equalcont{These authors contribute equally}

\author*[2]{\fnm{Wenbin} \sur{Xu}}\email{wenbinxu@lbl.gov}\equalcont{These authors contribute equally}

\author[3]{\fnm{Jie} \sur{Cao}}

\author*[4]{\fnm{Weilu} \sur{Gao}}\email{weilu.gao@utah.edu}

\author[2]{\fnm{Steven} \sur{Farrell}}

\author[5,6]{\fnm{Benjamin} \sur{Erichson}}

\author[5,6,7]{\fnm{Michael W.} \sur{Mahoney}}

\author[1]{\fnm{Andy} \sur{Nonaka}}

\author*[1]{\fnm{Zhi} \sur{Yao}}\email{jackie\_zhiyao@lbl.gov}

\affil[1]{Applied Mathematics and Computational Research Division, Lawrence Berkeley National Laboratory, Berkeley, CA, USA}

\affil[2]{National Energy Research Scientific Computing Center, Lawrence Berkeley National Laboratory, Berkeley, CA, USA}

\affil[3]{NSF National AI Institute for Student-AI Teaming, University of Colorado at Boulder, Boulder, USA}

\affil[4]{Department of Electrical and Computer Engineering, The University of Utah, Salt Lake City, UT, USA}

\affil[5]{Scientific Data Division, Lawrence Berkeley National Laboratory, Berkeley, CA, USA}

\affil[6]{International Computer Science Institute, Berkeley, CA, USA}

\affil[7]{Department of Statistics, University of California at Berkeley, Berkeley, CA, USA}

\keywords{Foundational model, Large Language Model, multi-modal learning, Material discovery}

%%\pacs[JEL Classification]{D8, H51}

%%\pacs[MSC Classification]{35A01, 65L10, 65L12, 65L20, 65L70}

\maketitle

\newpage
\begin{figure}[hbt]
    \centering
    \includegraphics[width = \textwidth]{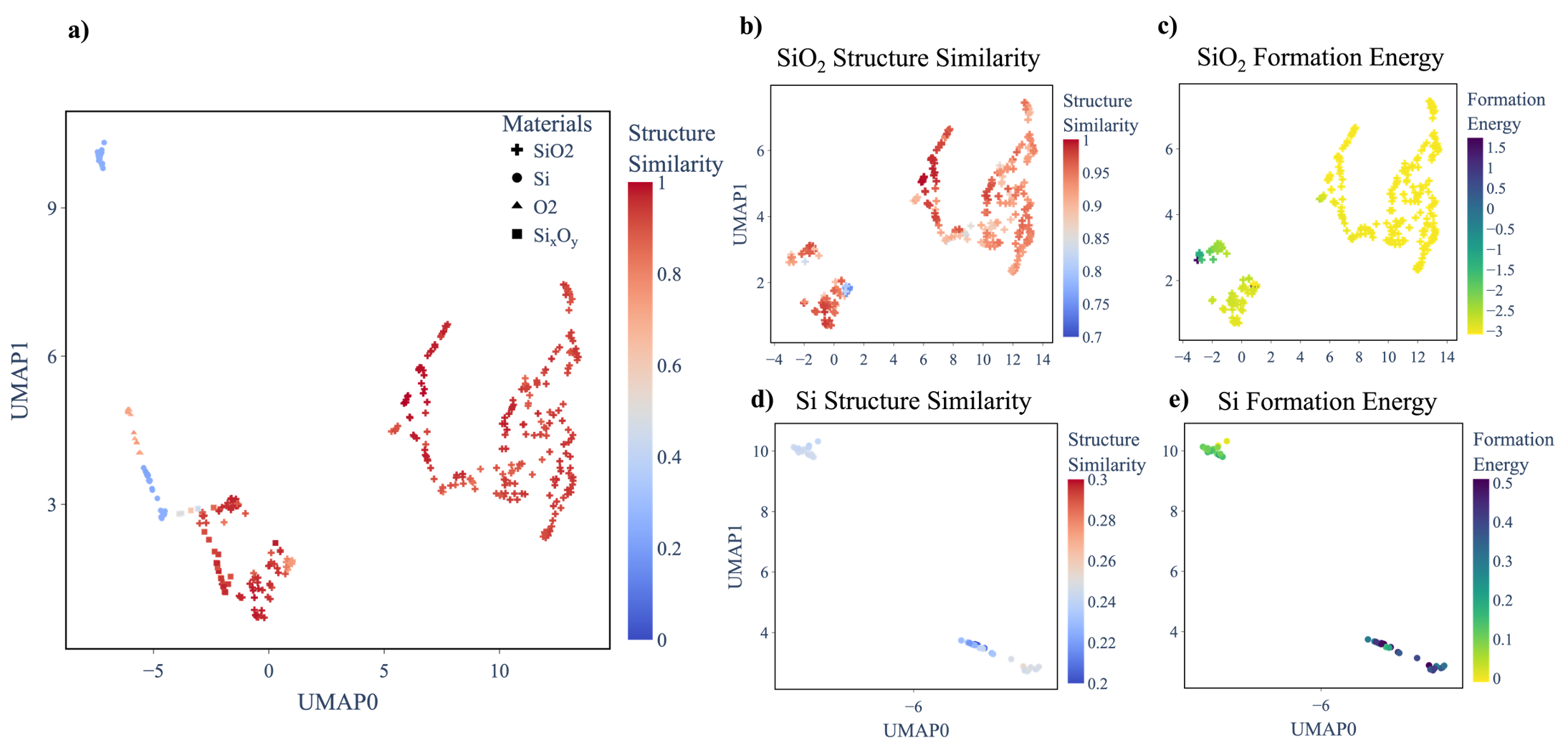}
    \caption{UMAP visualization of structural embeddings extracted from the bridge model. ($Si,$ $O$, $Si_xO_y$)}
    \label{SI_fig:xbar}
\end{figure}

\newpage
\begin{figure}[hbt]
    \centering
    \includegraphics[width = \textwidth]{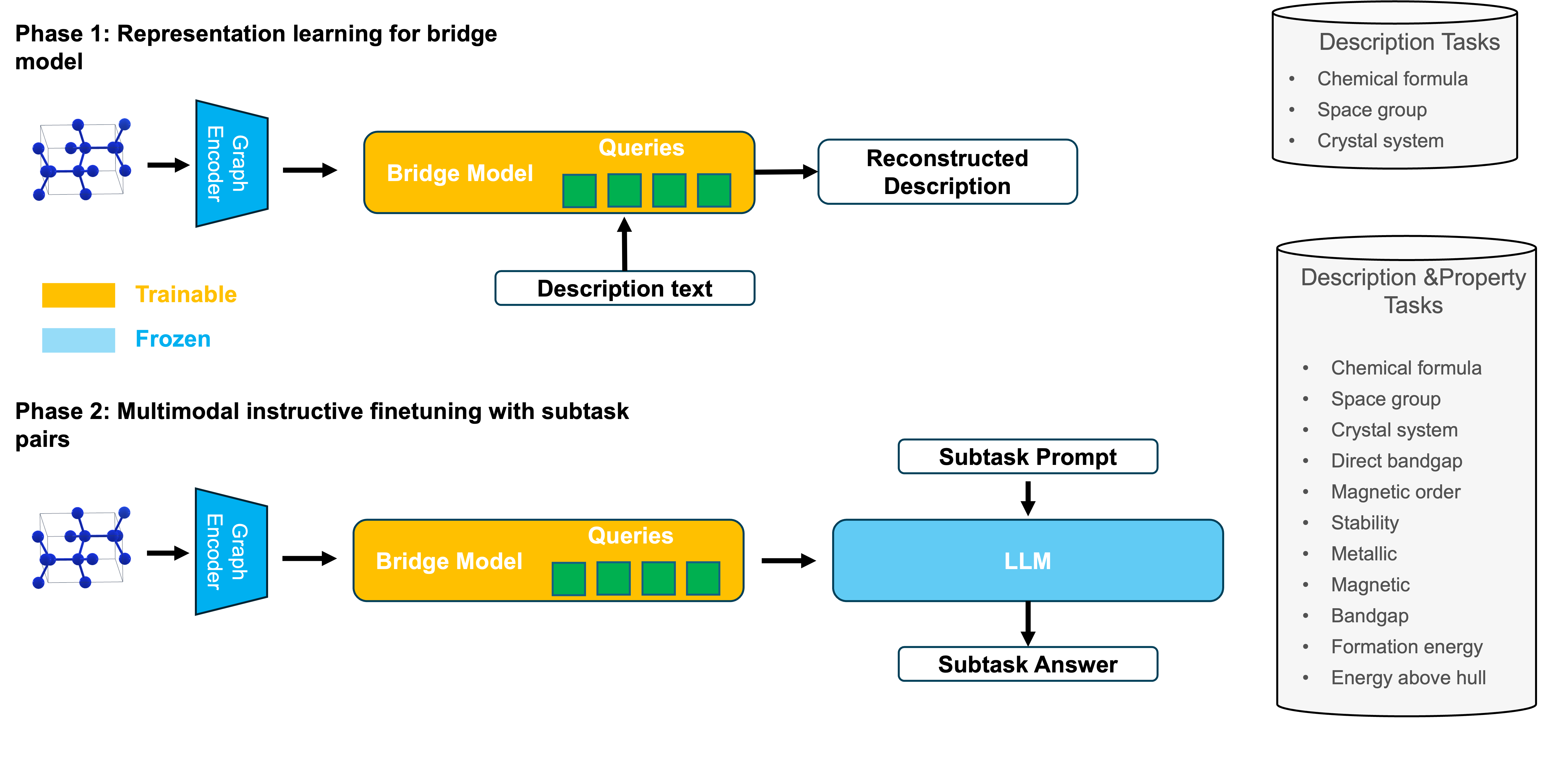}
    \caption{Training Scheme.}
    \label{SI_fig:xbar}
\end{figure}

\newpage
\section*{\quad Instruction Templates (descriptive tasks)}

\begin{table}[h!]
\centering
\renewcommand{\arraystretch}{1.2} % Adjust row height
\setlength{\tabcolsep}{5pt} % Adjust column separation
\begin{tabular}{|p{3cm}|p{12cm}|}
\hline
\textbf{Task} & \textbf{Instruction Template} \\
\hline
\textbf{Reduced Formula} & 
\begin{tabular}[t]{@{}l@{}}
\textless material structure\textgreater What is the chemical formula for this material? \\
\textless material structure\textgreater Can you tell me the chemical formula of this material? \\
\textless material structure\textgreater Please provide the chemical formula for the material. \\
\textless material structure\textgreater What is the formula for this material? \\
\textless material structure\textgreater Could you tell me the formula of the material? \\
\textless material structure\textgreater What elements make up this material? \\
\textless material structure\textgreater How would you write the chemical formula of this material? \\
\textless material structure\textgreater What is the exact chemical formula of this material? \\
\textless material structure\textgreater Can you provide the chemical formula for this material? \\
\end{tabular} \\
\hline
\textbf{Space Group} & 
\begin{tabular}[t]{@{}l@{}}
\textless material structure\textgreater What is the space group for this material? \\
\textless material structure\textgreater To which space group does this material belong? \\
\textless material structure\textgreater Can you tell me the space group of this material? \\
\textless material structure\textgreater Please provide the space group for the material. \\
\textless material structure\textgreater What is the crystallographic space group of this material? \\
\textless material structure\textgreater How is the space group of this material classified? \\
\textless material structure\textgreater Can you specify the space group for this material? \\
\textless material structure\textgreater Could you tell me the space group classification of this material? \\
\textless material structure\textgreater Can you provide the space group information for this material? \\
\textless material structure\textgreater What is the space group number of this material? \\
\end{tabular} \\
\hline
\textbf{Crystal System} & 
\begin{tabular}[t]{@{}l@{}}
\textless material structure\textgreater What is the crystal system of this material? \\
\textless material structure\textgreater Can you tell me the crystal system of this material? \\
\textless material structure\textgreater Please provide the crystal system for the material. \\
\textless material structure\textgreater What crystal system does this material belong to? \\
\textless material structure\textgreater How is the crystal system of this material classified? \\
\textless material structure\textgreater Can you specify the crystal system for this material? \\
\textless material structure\textgreater What is the crystallographic system of this material? \\
\textless material structure\textgreater Could you tell me the crystal system classification of this material? \\
\textless material structure\textgreater Which crystallographic system does this material belong to? \\
\end{tabular} \\
\hline
\textbf{Generate} & 
\begin{tabular}[t]{@{}l@{}}
\textless material structure\textgreater Can you provide another material similar to this material? \\
\textless material structure\textgreater Is there another material like this material that you can provide? \\
\textless material structure\textgreater Can you show me a different material similar to this one? \\
\textless material structure\textgreater Can you generate another material similar to this one? \\
\end{tabular} \\
\hline
\textbf{General} & 
\begin{tabular}[t]{@{}l@{}}
\textless material structure\textgreater Can you describe this material? \\
\textless material structure\textgreater \textless s\textgreater \\
\end{tabular} \\
\hline
\end{tabular}
\end{table}

\newpage
\section*{Answer Templates (descriptive tasks)}

\begin{table}[h!]
\centering
\renewcommand{\arraystretch}{1.2} % Adjust row height
\setlength{\tabcolsep}{5pt} % Adjust column separation
\begin{tabular}{|p{3cm}|p{12cm}|}
\hline
\textbf{Task} & \textbf{Instruction Template} \\
\hline
\textbf{Reduced Formula} & 
\begin{tabular}[t]{@{}l@{}}
The chemical formula for this material is \textless material attribute \textgreater. \\
The chemical formula of this material is \textless material attribute \textgreater. \\
The chemical formula for the material is \textless material attribute \textgreater. \\
The formula for this material is \textless material attribute \textgreater. \\
The formula of the material is \textless material attribute \textgreater. \\
The elements that make up this material are represented as \textless material attribute \textgreater. \\
The chemical formula of this material is written as \textless material attribute \textgreater. \\
The exact chemical formula of this material is \textless material attribute \textgreater. \\
The chemical formula for this material is \textless material attribute \textgreater. \\
\end{tabular} \\
\hline
\textbf{Space Group} & 
\begin{tabular}[t]{@{}l@{}}
The space group for this material is \textless material attribute \textgreater. \\
This material belongs to the space group \textless material attribute \textgreater. \\
The space group of this material is \textless material attribute \textgreater. \\
The space group for the material is \textless material attribute \textgreater. \\
The crystallographic space group of this material is \textless material attribute \textgreater. \\
The space group of this material is classified as \textless material attribute \textgreater. \\
The space group for this material is specified as \textless material attribute \textgreater. \\
The space group classification of this material is \textless material attribute \textgreater. \\
The space group information for this material is \textless material attribute \textgreater. \\
The space group number of this material is \textless material attribute \textgreater. \\
\end{tabular} \\
\hline
\textbf{Crystal System} & 
\begin{tabular}[t]{@{}l@{}}
The crystal system of this material is \textless material attribute \textgreater. \\
The crystal system of this material is \textless material attribute \textgreater. \\
The crystal system for the material is \textless material attribute \textgreater. \\
This material belongs to the \textless material attribute \textgreater crystal system. \\
The crystal system of this material is classified as \textless material attribute \textgreater. \\
The crystal system for this material is specified as \textless material attribute \textgreater. \\
The crystallographic system of this material is \textless material attribute \textgreater. \\
The crystal system classification of this material is \textless material attribute \textgreater. \\
This material belongs to the \textless material attribute \textgreater crystallographic system. \\
\end{tabular} \\
\hline
\end{tabular}
\end{table}

\newpage
\section*{\quad Instruction Templates (property part1)}

\begin{table}[h!]
\centering
\renewcommand{\arraystretch}{1.2} % Adjust row height
\setlength{\tabcolsep}{5pt} % Adjust column separation
\begin{tabular}{|p{3cm}|p{12cm}|}
\hline
\textbf{Task} & \textbf{Instruction Template} \\
\hline
\textbf{Is Metal} & 
\begin{tabular}[t]{@{}l@{}}
\textless material structure\textgreater Is this material metal or non-metal? \\
\textless material structure\textgreater Can you tell me if this material is metal or not? \\
\textless material structure\textgreater What is the classification of this material: metal or non-metal? \\
\textless material structure\textgreater Is this material considered a metal? \\
\textless material structure\textgreater How is this material categorized: metal or non-metal? \\
\textless material structure\textgreater Could you specify if this material is metal or non-metal? \\
\textless material structure\textgreater Is the material metallic or non-metallic? \\
\textless material structure\textgreater Can you provide the classification of this material: metal or non-metal? \\
\textless material structure\textgreater Is this material identified as a metal or non-metal? \\
\textless material structure\textgreater What type of material is this: metal or non-metal? \\
\end{tabular} \\
\hline
\textbf{Direct Bandgap} & 
\begin{tabular}[t]{@{}l@{}}
\textless material structure\textgreater Does the material have a direct bandgap or indirect bandgap? \\
\textless material structure\textgreater Is the bandgap of this material direct or indirect? \\
\textless material structure\textgreater Can you tell me if this material has a direct or indirect bandgap? \\
\textless material structure\textgreater What type of bandgap does this material have: direct or indirect? \\
\textless material structure\textgreater Is this material characterized by a direct or indirect bandgap? \\
\textless material structure\textgreater Could you specify if the bandgap of this material is direct or indirect? \\
\textless material structure\textgreater Does this material exhibit a direct or indirect bandgap? \\
\textless material structure\textgreater Is the bandgap in this material direct or indirect? \\
\textless material structure\textgreater How is the bandgap of this material classified: direct or indirect? \\
\textless material structure\textgreater Is this a direct or indirect bandgap material? \\
\end{tabular} \\
\hline
\textbf{Stability} & 
\begin{tabular}[t]{@{}l@{}}
\textless material structure\textgreater Is this material stable? \\
\textless material structure\textgreater Can you tell me if this material is stable? \\
\textless material structure\textgreater What is the stability of this material? \\
\textless material structure\textgreater Please provide the stability information for this material. \\
\textless material structure\textgreater Is the material stable under standard conditions? \\
\textless material structure\textgreater Is this material thermodynamically stable? \\
\end{tabular} \\
\hline
\textbf{Experimental Observation} & 
\begin{tabular}[t]{@{}l@{}}
\textless material structure\textgreater Is the material experimentally observed or not? \\
\textless material structure\textgreater Can you tell me if the material is observed in experiments? \\
\end{tabular} \\
\hline
\textbf{Is Magnetic} & 
\begin{tabular}[t]{@{}l@{}}
\textless material structure\textgreater Is the material magnetic or not? \\
\textless material structure\textgreater Is the material magnetic or non-magnetic? \\
\textless material structure\textgreater Can you tell me if this material is magnetic? \\
\textless material structure\textgreater What is the magnetic nature of this material? \\
\textless material structure\textgreater Is this material classified as magnetic? \\
\textless material structure\textgreater Does this material have magnetic properties? \\
\textless material structure\textgreater Is this a magnetic or non-magnetic material? \\
\end{tabular} \\
\hline
\textbf{Magnetic Order} & 
\begin{tabular}[t]{@{}l@{}}
\textless material structure\textgreater What is the magnetic order of the material? \\
\textless material structure\textgreater Can you tell me the magnetic order of this material? \\
\textless material structure\textgreater Could you specify the magnetic order of the material? \\
\textless material structure\textgreater What type of magnetic order does this material have? \\
\textless material structure\textgreater Please provide the magnetic ordering of the material. \\
\textless material structure\textgreater What is the magnetic arrangement in this material? \\
\textless material structure\textgreater Could you tell me the type of magnetic order of this material? \\
\end{tabular} \\
\hline
\end{tabular}
\end{table}

\newpage
\section*{Answer Templates (property part1)}

\begin{table}[h!]
\centering
\renewcommand{\arraystretch}{1.2} % Adjust row height
\setlength{\tabcolsep}{5pt} % Adjust column separation
\begin{tabular}{|p{3cm}|p{11cm}|}
\hline
\textbf{Task} & \textbf{Instruction Template} \\
\hline
\textbf{Is Metal} & 
\begin{tabular}[t]{@{}l@{}}
This material is classified as \textless material attribute \textgreater. \\
This material is a \textless material attribute \textgreater. \\
The classification of this material is \textless material attribute \textgreater. \\
This material is considered \textless material attribute \textgreater. \\
This material is categorized as \textless material attribute \textgreater. \\
This material is specified as \textless material attribute \textgreater. \\
This material is \textless material attribute \textgreater. \\
The classification of this material is \textless material attribute \textgreater. \\
This material is identified as \textless material attribute \textgreater. \\
This type of material is \textless material attribute \textgreater. \\
\end{tabular} \\
\hline
\textbf{Direct Bandgap} & 
\begin{tabular}[t]{@{}l@{}}
The material has a \textless material attribute \textgreater bandgap. \\
The bandgap of this material is \textless material attribute \textgreater. \\
This material has a \textless material attribute \textgreater bandgap. \\
This material has a \textless material attribute \textgreater type of bandgap. \\
This material is characterized by a \textless material attribute \textgreater bandgap. \\
The bandgap of this material is specified as \textless material attribute \textgreater. \\
This material exhibits a \textless material attribute \textgreater bandgap. \\
The bandgap in this material is \textless material attribute \textgreater. \\
The bandgap of this material is classified as \textless material attribute \textgreater. \\
This is a \textless material attribute \textgreater bandgap material. \\
\end{tabular} \\
\hline
\textbf{Stability} & 
\begin{tabular}[t]{@{}l@{}}
This material is \textless material attribute \textgreater. \\
Yes, this material is \textless material attribute \textgreater. \\
The stability of this material is \textless material attribute \textgreater. \\
The stability information for this material is \textless material attribute \textgreater. \\
This material is \textless material attribute \textgreater under standard conditions. \\
This material is \textless material attribute \textgreater. \\
\end{tabular} \\
\hline
\textbf{Experimental Observation} & 
\begin{tabular}[t]{@{}l@{}}
The material is \textless material attribute \textgreater. \\
The material is \textless material attribute \textgreater. \\
\end{tabular} \\
\hline
\textbf{Is Magnetic} & 
\begin{tabular}[t]{@{}l@{}}
The material is \textless material attribute \textgreater. \\
This material is \textless material attribute \textgreater. \\
Yes, this material is \textless material attribute \textgreater. \\
The magnetic nature of this material is \textless material attribute \textgreater. \\
This material is classified as \textless material attribute \textgreater. \\
This material has \textless material attribute \textgreater properties. \\
This is a \textless material attribute \textgreater material. \\
\end{tabular} \\
\hline
\textbf{Magnetic Order} & 
\begin{tabular}[t]{@{}l@{}}
The magnetic order of the material is \textless material attribute \textgreater. \\
The magnetic order of this material is \textless material attribute \textgreater. \\
The magnetic order of the material is specified as \textless material attribute \textgreater. \\
This material has a \textless material attribute \textgreater type of magnetic order. \\
The magnetic ordering of the material is \textless material attribute \textgreater. \\
The magnetic arrangement in this material is \textless material attribute \textgreater. \\
The type of magnetic order of this material is \textless material attribute \textgreater. \\
\end{tabular} \\
\hline
\end{tabular}
\end{table}

\newpage
\section*{\quad Instruction Templates (property part2)}

\begin{table}[h!]
\centering
\renewcommand{\arraystretch}{1.2} % Adjust row height
\setlength{\tabcolsep}{5pt} % Adjust column separation
\begin{tabular}{|p{3cm}|p{12cm}|}
\hline
\textbf{Task} & \textbf{Instruction Template} \\
\hline
\textbf{Bandgap} & 
\begin{tabular}[t]{@{}l@{}}
\textless material structure\textgreater What is the bandgap of the material? \\
\textless material structure\textgreater Can you tell me the bandgap of this material? \\
\textless material structure\textgreater What is the energy bandgap for this material? \\
\textless material structure\textgreater Could you specify the bandgap of the material? \\
\textless material structure\textgreater Could you tell me the bandgap energy level of this material? \\
\end{tabular} \\
\hline
\textbf{Formation Energy} & 
\begin{tabular}[t]{@{}l@{}}
\textless material structure\textgreater Can you tell me the formation energy of this material? \\
\textless material structure\textgreater Please provide the formation energy for the material. \\
\textless material structure\textgreater What is the formation energy value for this material? \\
\textless material structure\textgreater How much is the formation energy of this material? \\
\textless material structure\textgreater Can you specify the formation energy of this material? \\
\end{tabular} \\
\hline
\textbf{Energy Above Hull} & 
\begin{tabular}[t]{@{}l@{}}
\textless material structure\textgreater Can you tell me the energy above hull of this material? \\
\textless material structure\textgreater Please provide the energy above hull for the material. \\
\textless material structure\textgreater What is the energy above the hull for this material? \\
\textless material structure\textgreater How much is the energy above hull for this material? \\
\textless material structure\textgreater Can you specify the energy above hull of this material? \\
\textless material structure\textgreater Could you tell me the energy above hull of the material? \\
\end{tabular} \\
\hline
\end{tabular}
\end{table}

\newpage
\section*{Answer Templates (property part2)}

\begin{table}[h!]
\centering
\renewcommand{\arraystretch}{1.2} % Adjust row height
\setlength{\tabcolsep}{5pt} % Adjust column separation
\begin{tabular}{|p{3cm}|p{12cm}|}
\hline
\textbf{Task} & \textbf{Instruction Template} \\
\hline
\textbf{Bandgap} & 
\begin{tabular}[t]{@{}l@{}}
The bandgap of the material is \textless material attribute \textgreater. \\
The bandgap of this material is \textless material attribute \textgreater. \\
The energy bandgap for this material is \textless material attribute \textgreater. \\
The bandgap of the material is specified as \textless material attribute \textgreater. \\
The bandgap energy level of this material is \textless material attribute \textgreater. \\
\end{tabular} \\
\hline
\textbf{Formation Energy} & 
\begin{tabular}[t]{@{}l@{}}
The formation energy of this material is \textless material attribute \textgreater. \\
The formation energy for the material is \textless material attribute \textgreater. \\
The formation energy value for this material is \textless material attribute \textgreater. \\
The formation energy of this material is \textless material attribute \textgreater. \\
The formation energy of this material is specified as \textless material attribute \textgreater. \\
\end{tabular} \\
\hline
\textbf{Energy Above Hull} & 
\begin{tabular}[t]{@{}l@{}}
The energy above hull of this material is \textless material attribute \textgreater. \\
The energy above hull for the material is \textless material attribute \textgreater. \\
The energy above the hull for this material is \textless material attribute \textgreater. \\
The energy above hull for this material is \textless material attribute \textgreater. \\
The energy above hull of this material is specified as \textless material attribute \textgreater. \\
The energy above hull of the material is \textless material attribute \textgreater. \\
\end{tabular} \\
\hline
\end{tabular}
\end{table}